\documentclass{article} %
\usepackage{iclr2024_conference,times}

\usepackage{amsmath,amsfonts,bm}

\def\eqref#1{equation~\ref{#1}}

\def\1{\bm{1}}

\DeclareMathAlphabet{\mathsfit}{\encodingdefault}{\sfdefault}{m}{sl}
\SetMathAlphabet{\mathsfit}{bold}{\encodingdefault}{\sfdefault}{bx}{n}

\usepackage{graphicx}
\usepackage[pagebackref=true,breaklinks=true,letterpaper=true,colorlinks,bookmarks=false,citecolor=cyan]{hyperref}

\usepackage{url}
\usepackage{wrapfig}
\usepackage{amsfonts}       %
\usepackage{booktabs}
\usepackage{multirow}
\usepackage{authblk}
\usepackage{makecell}

\usepackage{xspace}
\def\Ours{Uni3D\xspace}

\newcommand{\authorskip}{\hspace{12mm}}

\begin{document}

\title{\Ours: Exploring Unified 3D Representation at Scale}

\author{Junsheng Zhou\textsuperscript{1,2}\thanks{Equal contribution. Correspondence to \textit{\{brma@baai.ac.cn\}} and \textit{\{wangxinlong@baai.ac.cn\}}. 
} 
\authorskip Jinsheng Wang\textsuperscript{1}$^*$ \authorskip Baorui Ma\textsuperscript{1}$^*$ \authorskip Yu-Shen Liu\textsuperscript{2} \authorskip  
Tiejun Huang\textsuperscript{1,3} \authorskip Xinlong Wang\textsuperscript{1}  \\
{
\fontsize{10.4pt}{9.84pt}\selectfont
\textsuperscript{1} Beijing Academy of Artificial Intelligence \hspace{5mm} 
\textsuperscript{2} Tsinghua University \hspace{5mm} \textsuperscript{3} Peking University}\\

{
\fontsize{9.4pt}{9.84pt}\selectfont 
\vspace{0.2cm}
Code \& Models: {\url{https://github.com/baaivision/Uni3D}}
}
}

\newcommand{\fix}{\marginpar{FIX}}
\newcommand{\new}{\marginpar{NEW}}

\iclrfinalcopy %

\maketitle

\begin{abstract}

Scaling up representations for images or text has been extensively investigated in the past few years and has led to revolutions in learning vision and language.
However, scalable representation for 3D objects and scenes is relatively unexplored.
In this work, we present \Ours, a 3D foundation model to explore the unified 3D representation at scale.
\Ours uses a 2D initialized ViT end-to-end pretrained to align the 3D point cloud features with the image-text aligned features.
Via the simple architecture and pretext task, \Ours can leverage abundant 2D pretrained models as initialization and image-text aligned models as the target, unlocking the great potential of 2D models and scaling-up strategies to the 3D world.
We efficiently scale up \Ours to one billion parameters, and set new records on a broad range of 3D tasks, such as zero-shot classification, few-shot classification, open-world understanding and part segmentation. 
We show that the strong \Ours representation also enables applications such as 3D painting and retrieval in the wild.
We believe that \Ours provides a new direction for exploring both scaling up and efficiency of the representation in 3D domain.

\end{abstract}

\section{Introduction}

3D representation learning is one of the most fundamental problems in 3D computer vision, especially with the rapid development of 3D sensors (e.g., LiDAR) and the growing demands in real-world applications, e.g., autonomous driving, augmented/virtual reality and robotics.  
Existing methods make great progress in 3D model architecture~\citep{qi2017pointnet, qi2017pointnet++, yu2021pointr, wang2019dynamic}, learning objective~\citep{yu2022point, wang2021unsupervised}, task-oriented modeling~\citep{zhou2020cylinder3d, yin2021center, zhao2021point}, etc.
However, most of the works explore at a relatively small scale, with limited parameters, data, and task scenarios. 
Learning scalable 3D representation that can transfer in the wild is relatively unexplored and remains a challenging problem.

\begin{figure}[tb]
\begin{minipage}{0.53\textwidth}
  \centering
  \includegraphics[width=\linewidth]{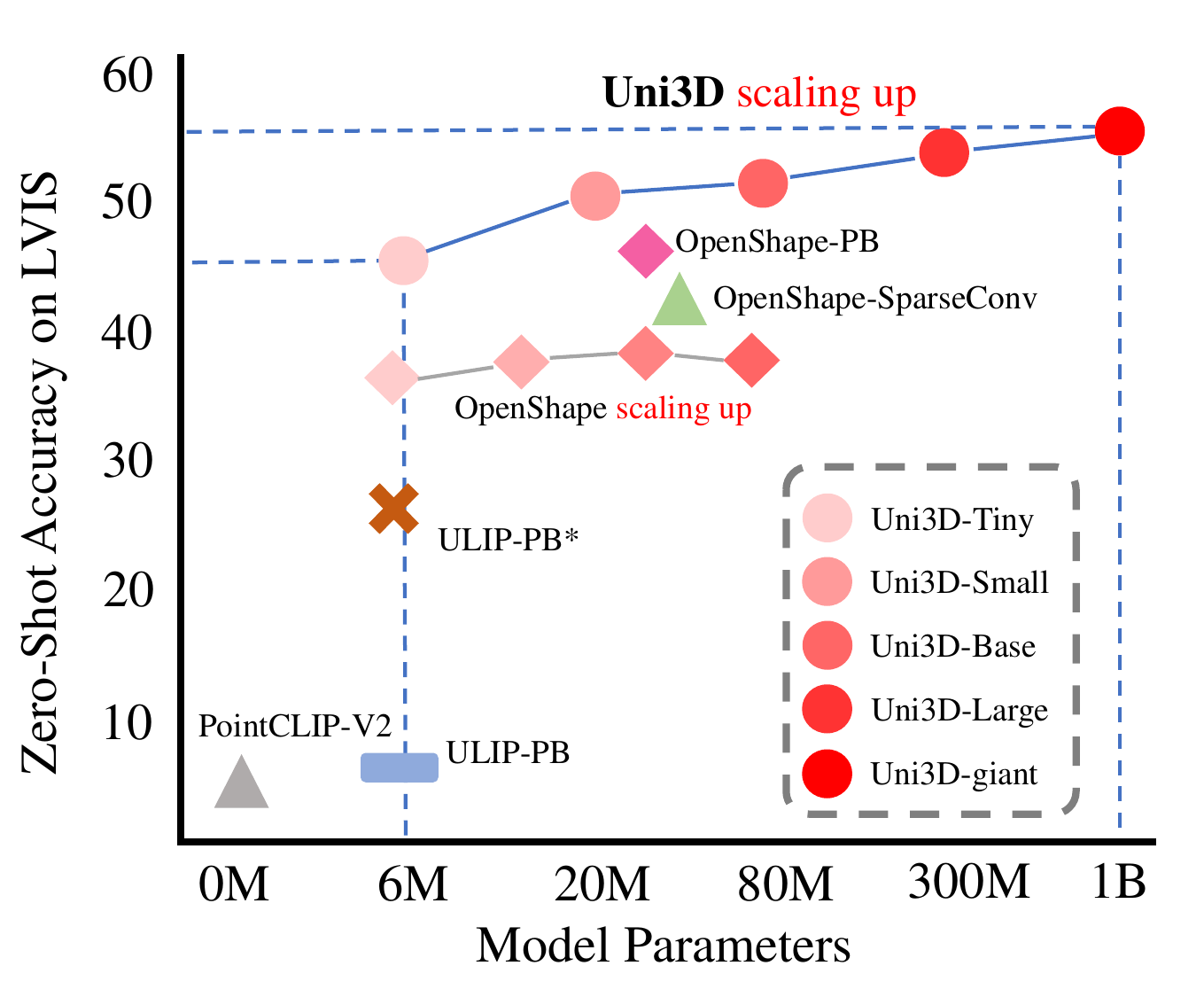}
  \vspace{-0.7cm}
  \caption{The parameter and zero-shot accuracy comparison. Uni3D scales up 3D representation from 6M to 1B. `PB' indicates PointBERT. `ULIP-PB*' indicates retrained ULIP on the ensambled large dataset.}
  \label{fig:scaling}
\end{minipage}%
\hspace{0.2cm}
\begin{minipage}{0.44\textwidth}
  \centering
  \includegraphics[width=\linewidth]{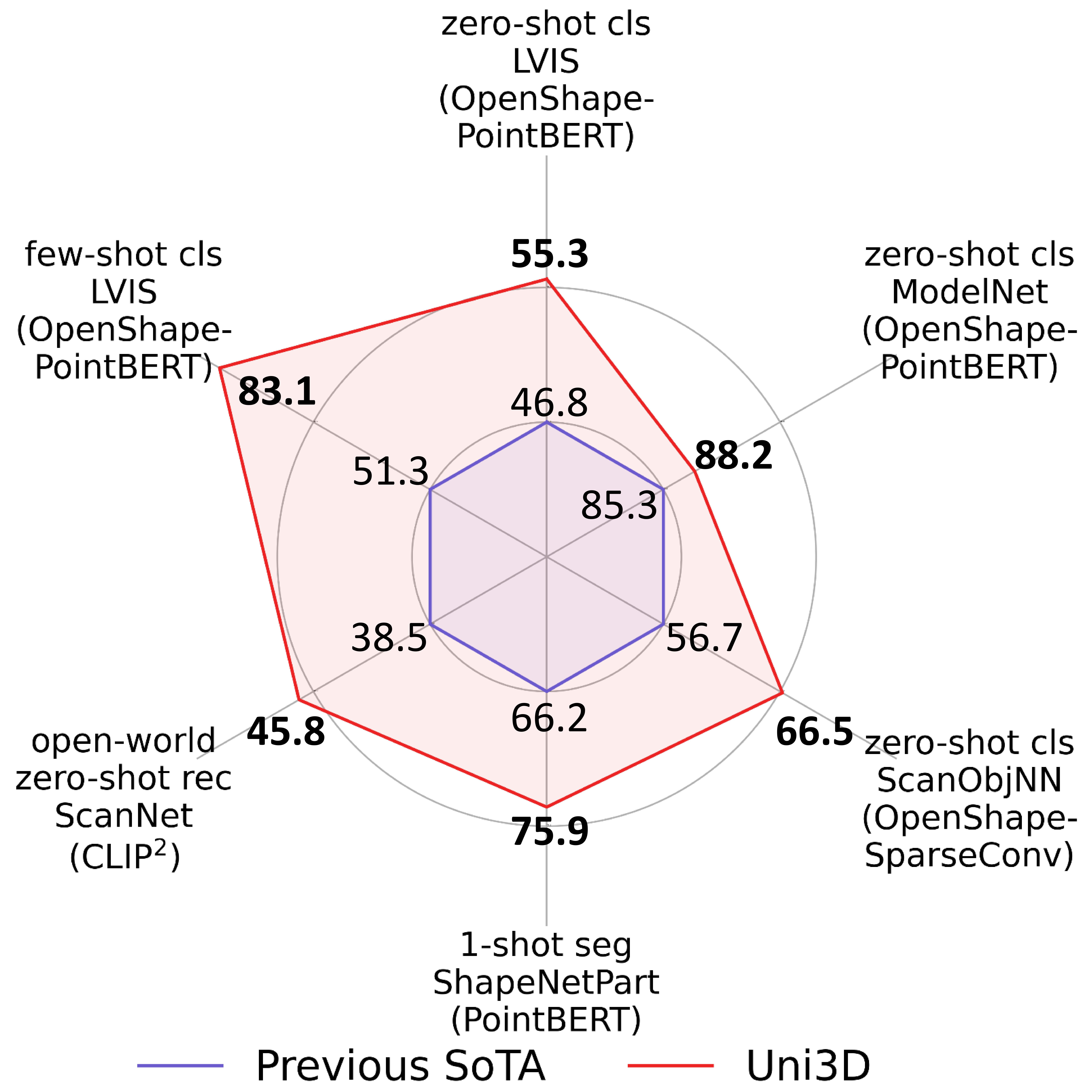}
  \vspace{-0.6cm}
  \caption{Qualitative comparisons between \Ours and previous SoTA methods under different tasks and benchmarks. The scale of each axis in the radar chart is normalized by the performance of \Ours .}
  \label{fig:radar}
\end{minipage}
\vspace{-0.45cm}
\end{figure}

In the past few years, scaling up pre-trained language models~\citep{brown2020language, liu2019roberta, raffel2020exploring} has largely revolutionized natural language processing.
Some recent works~\citep{radford2021learning, dosovitskiy2020image, bao2021beit, he2022masked, fang2023eva} translate the progress from language to 2D vision via model and data scaling. 
Motivated by their success, it is appealing that we can also lift this success from 2D to 3D, i.e., to learn a scalable 3D representation model that can transfer in the 3D world.
Recently, as the release of a large-scale 3D dataset Objaverse~\citep{deitke2023objaverse}, a few works have tried to explore scalable pretraining in 3D, but either still limit to the small-scale 3D backbones~\citep{xue2023ulip, xue2023ulip2}, or can hardly scale to a relatively larger size~\citep{liu2023openshape}, e.g., 72M in Fig.~\ref{fig:scaling}.

In this work, we propose \Ours, a unified and scalable 3D pretraining framework for large-scale 3D representation learning, and explore its limits at the scale of one billion parameters with a million 3D shapes and 10 million images paired with 70 million texts. 
\Ours uses a 2D ViT as the 3D encoder initialized with the best 2D prior, which is then end-to-end pre-trained to align the 3D point cloud features with the image-text aligned features.
Via the simple architecture and pretext task, \Ours can leverage abundant 2D pre-trained models as initialization~\citep{fang2023eva,  caron2021emerging},  and image-text aligned models as the target~\citep{radford2021learning, sun2023eva, cherti2023reproducible}, unlocking the great potential of 2D models and scaling-up strategies to the 3D world.

In addition, we systematically study the scalability and flexibility of \Ours in terms of 1) model scaling from 6M to 1B parameters, 2) 2D initialization from visual self-supervised to text supervised, and 3) text-image aligned target model from 150M to 5B parameters.
We observe continuous performance improvements as the scaling of each component under the flexible and unified framework.
The sharable 2D prior and scale-up strategies also largely benefit the large-scale 3D representation learning.

For the first time, we demonstrate a billion-scale 3D representation model that transfers well to various downstream tasks and scenarios.
As shown in Fig. \ref{fig:radar}, \Ours yields a boost compared to prior art in various zero-shot and few-shot 3D tasks.
Specifically, \Ours achieves a zero-shot classification accuracy of 88.2\% on ModelNet, which surprisingly performs on par with some supervision methods. 
\Ours also achieves state-of-the-art performance on other representative 3D tasks such as open-world understanding, part segmentation, etc.
In addition, we present some interesting applications with the strong 3D representation learned by \Ours, such as point cloud painting and text/image-based  3D shape retrieval. 

By scaling up 3D foundation models with a simple and unified pre-training to learn strong 3D representation across tasks, we hope \Ours would bridge the gap between 2D and 3D vision, and contribute to the big convergence across different modalities. To facilitate future research, we will release all the code and 3D foundation models.

\section{Related Work}

\textbf{3D Representation Learning.}
Learning representations from point clouds for 3D understanding \citep{qi2017pointnet, qi2017pointnet++, wang2019dynamic, yu2021pointr} has been fully explored in recent years. Some works further studied self-supervised pretraining for point clouds by specific 3D pretext tasks like self-reconstruction \citep{wang2021unsupervised}, mask point modeling \citep{yu2022point} and contrastive learning \citep{qi2023contrast}. These works merely explore under limited 3D data (e.g. ShapeNet \citep{chang2015shapenet}) and do not investigate multi-modal representation from 2D/NLP to 3D.

With the recent success in learning visual concepts from raw text with contrastive learning like CLIP ~\citep{radford2021learning,jia2021scaling,li2022grounded,ramesh2022hierarchical,gregoromichelaki2022language},
recent works ~\citep{ liu2023openshape, qi2023contrast,xue2023ulip,hegde2023clip, lei2023vit} seek to learn 3D representations by aligning text, image, and point cloud features through in a similar contrastive learning way. Recently, as the release of a large-scale 3D dataset Objaverse \citep{deitke2023objaverse}, OpenShape~\citep{ liu2023openshape} and ULIP2~\citep{xue2023ulip2} have tried to explore scalable pretraining in 3D, but either still limit to the small-scale 3D backbones \citep{xue2023ulip2}, or can hardly scale to a relatively larger size \citep{liu2023openshape}. In this work, We aim to explore a unified and scalable 3D pretraining framework, i.e., \Ours, for large-scale 3D representation learning and explore its limits in billion-scale model sizes.

\textbf{Foundation models.}
Recently, it has been drawing significant attention to design foundation models for unifying and scaling up representations under different modalities (e.g. NLP, 2D vision). Starting from NLP, recent works in scaling up pre-trained language models~\citep{brown2020language, liu2019roberta, raffel2020exploring} have largely revolutionized natural language processing.
Some research in 2D vision~\citep{radford2021learning, dosovitskiy2020image, bao2021beit, he2022masked, fang2023eva} translates the progress from language to 2D vision via model and data scaling.  However, such a phenomenon has not been well-established and explored in the 3D domain, due to the limited 3D data and difficulties in unifying and scaling up 3D backbones. 
Meta-Transformer ~\citep{zhang2023metatransformer} and FrozeCLIP~\citep{huang2022frozen} have indicated a promising future for developing a unified framework with a modality-shared encoder. However, they require retraining task-specific heads with labor-intensive manual labeling of ground truth for different downstream tasks, which leads to a lack of out-of-domain zero-shot capabilities. 
In this work, we design the first billion-scale 3D foundation model with a unified 3D representation. 
The unified ViT architecture allows us to simply scale up \Ours with the well-studied unified 2D/NLP scaling-up strategies. 
We anticipate \Ours to serve as a bridge between 2D and 3D vision, facilitating significant convergence across various modalities.

\section{Method}

\begin{figure}[tb]
    \centering
    \includegraphics[width=1\textwidth]{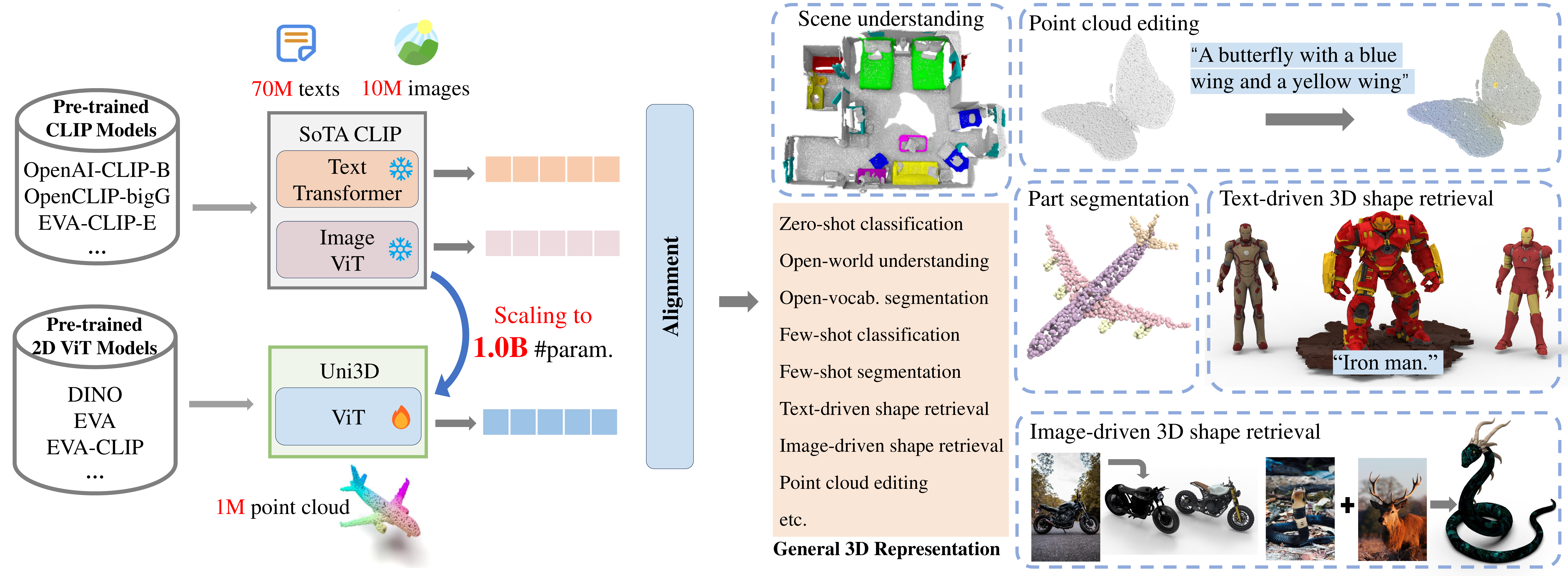}
    \caption{\textbf{The overview of \Ours}. \Ours is a unified and scalable 3D pretraining framework for large-scale 3D representation learning. We scale up \Ours to one billion parameters with a million 3D shapes paired with 10 million images and 70 million texts. Uni3D uses a 2D ViT as the 3D encoder initialized with the best 2D prior from abundant 2D pre-trained models, which is then end-to-end pre-trained to align the 3D point cloud features with the image-text aligned ones from SoTA CLIP models. \Ours shows superior performance on a wide range of benchmarks.}
    \label{fig:overview}
    \vspace{-0.3cm}
\end{figure}

We introduce \Ours, a unified and scalable 3D pretraining framework for large-scale 3D representation learning by aligning 3D point cloud features with the image-text aligned features. The overview of \Ours is shown in Fig. \ref{fig:overview}. We first present how we design, scale up and initialize a unified 3D representation in \Ours in Sec. \ref{sec.m1}. We then introduce the multi-modal contrastive learning for aligning image and language with point cloud in Sec. \ref{sec.m2}. More training details are provided in Sec.~\ref{sec.m4} of the appendix.

\subsection{Unified 3D representation}
\label{sec.m1}
\Ours leverages a unified vanilla transformer structurally equivalent to 2D Vision Transformer (ViT) \citep{dosovitskiy2020image} as the backbone. The only difference here is that we replace the patch embedding layer in ViT with a specific point tokenizer to achieve 3D embeddings. The point tokenizer keeps the same as PointBERT \citep{yu2022point} to first group points into local patches with FPS (farthest point sampling) and kNN (k nearest neighbor), and then extract token embeddings with a tiny PointNet \citep{qi2017pointnet} for each patch. The vanilla transformer is then applied to the 3D tokens to extract the 3D representations. 

\noindent\textbf{Scaling Up \Ours.} 
Previous works on point cloud representation learning merely focus on designing specific model architectures for pursuing better performances in different applications and are limited to a certain small-scale dataset (e.g. ShapeNet \citep{chang2015shapenet}, ModelNet \citep{wu20153d}). 
With the recent successes in large-scale 3D data (e.g. Objaverse \citep{deitke2023objaverse,deitke2023objaversexl}), a few recent works \citep{xue2023ulip, liu2023openshape, xue2023ulip2} have tried to explore scalable pretraining in 3D, but either still limit to the small-scale 3D backbones \citep{xue2023ulip}, or can hardly scale to a relatively larger size \citep{liu2023openshape}. 
The difficulties lie in the un-unified backbones and pretraining in 3D domain, where each backbone requires a specific scaling-up strategy, which is rarely explored. 
Moreover, some backbones (e.g. PointMLP \citep{ma2021rethinking}, DGCNN \citep{wang2019dynamic}) require modeling local patterns completely on dense points, which brings extensive computational cost when scaling up.

We justify that \Ours, which directly leverages the vanilla transformer structurally equivalent to ViT, can naturally solve the difficulties by simply scaling up the model size with the well-studied unified 2D/NLP scaling-up strategies. Specifically, we leverage the strategy of ViT which gradually scales up Transformer from Tiny (6 M), Small (23M), Base (88 M), Large (307 M) to giant (1B) and replace the Transformer of \Ours with different sizes of ViT as the scaled-up version of \Ours at different model sizes. The effectiveness and efficiency of our scaling-up strategy are fully demonstrated by the comprehensive exploration of scaling up ViT in the 2D vision domain. 
As shown in Fig. \ref{fig:scaling} and Tab. \ref{tab:model_scaling}, we observe continuous performance improvements as the scaling of model size under the flexible and unified framework.

Given the unified scaling-up strategy, we train the largest 3D presentation model with one billion parameters under the multi-modal alignment learning objective, in a large-scale dataset of nearly one million 3D shapes, along with paired 10 million images and 70 million texts. For the first time, we demonstrate a billion-scale 3D representation model that transfers well to various downstream tasks and scenarios.

\noindent\textbf{Initializing \Ours.} Another challenge that prevents previous works in scaling up 3D backbones is that larger model sizes lead to overfitting and difficulties in convergence. A naive solution is to pretrain each 3D backbone with specific 3D pretext tasks (e.g. PointBERT \citep{yu2022point}, OcCo \citep{wang2021unsupervised}), and leverage the pretrained parameters as the initialization. 
However, this results in expensive training costs, and the relatively limited scale of 3D data for pretraining makes it challenging to establish a robust initialization for stabilizing cross-modal contrastive learning.

In \Ours, we directly leverage the vanilla transformer structurally equivalent to ViT as the 3D backbone, which brings a new perspective of introducing pretrained priors. Specifically, we can naturally adopt the pretrained large models in other modalities which share the same vanilla transformer as ours to initialize \Ours, such as the 2D pretrained model DINO \citep{caron2021emerging}, EVA \citep{fang2023eva}, EVA-02 \citep{fang2023eva02} and the cross-modal models CLIP \citep{radford2021learning}, EVA-CLIP \citep{sun2023eva}, etc. These pretrained models are trained in datasets consisting of billions of images and texts, which already learn rich underlying representational abilities for Transformer and have the potential to enhance and stabilize the learning of large-scale 3D representations. 
\Ours is not limited to a specific pretrained model for initialization, where we can flexibly leverage any off-the-shelf Transformer-based pretrained models at any modalities for pushing the performance and exploring the cross-modal pretraining (please refer to Sec. \ref{sec.init} for detailed analysis).

\subsection{Multi-Modal Alignment}
\label{sec.m2}

We train \Ours to learn the multi-modal alignment across language, image and point cloud following a similar paradigm as ULIP \citep{xue2023ulip} and OpenShape \citep{liu2023openshape}. 

\textbf{Datasets.} In order to keep the experimental settings consistent with other methods for a fair comparison, we adopt the ensembled 3D dataset provided by OpenShape for training, which consists of four 3D dataset, i.e., Objaverse \citep{deitke2023objaverse}, ShapeNet \citep{chang2015shapenet}, 3D-FUTURE \citep{fu20213d} and ABO \citep{collins2022abo}. We sample 10,000 points from the mesh surface with colors and render 10 color images from different views that uniformly cover the whole shape. The point cloud-text-image triplets are conducted in the same way as OpenShape.

\textbf{Objective.} The illustration of the multi-modal alignment is shown in Fig. \ref{fig:overview}. We initialize the \Ours point encoder $f_P$ with pretrained 2D ViT models and obtain the text encoder $f_T$ and image encoder $f_I$ from CLIP models. We train $f_P$ to learn 3D representations by aligning them to well-learned 2D / Language representations of CLIP models and distills cross-modal knowledge. Both $f_I$ and $f_T$ are frozen since they are well-optimized, and only $f_P$ are learnable during training. Given a batch of $N$ triplets $\{(P_i, I_i, T_i)\}_{i=1}^N$, where $P_i$, $I_i$ , $T_i$ donate a point cloud and its corresponding image and text obtained from the same 3D shape. We first achieve the normalized feature for the sampled triplets as $\{(e_i^P=f_P(P_i)/|f_P(P_i)|, e_i^I=f_I(I_i)/|f_I(P_i)|, e_i^T=f_T(T_i)/|f_T(T_i)|)\}_{i=1}^N$. The contrastive loss is then formulated as:

\begin{equation}
\scriptsize
-\frac{1}{4N} \sum_{i=1}^N\left(\log\frac{\exp(e_i^P\cdot e_i^T/\tau)}{\sum_{j} \exp(e_i^P\cdot e_j^T/\tau)} +
\log\frac{\exp(e_i^T\cdot e_i^P/\tau)}{\sum_{j} \exp(e_i^T\cdot e_j^P/\tau)} + 
\log\frac{\exp(e_i^P\cdot e_i^I/\tau)}{\sum_{j} \exp(e_i^P\cdot e_j^I/\tau)} +
\log\frac{\exp(e_i^I\cdot e_i^P/\tau)}{\sum_{j} \exp(e_i^I\cdot e_j^P/\tau)}\right),
\end{equation}

where $\tau$ is a learnable temperature. The training target is to minimize the triplet contrastive loss.

\textbf{Image-Text Aligned Target.} We further justify that \Ours is not limited to a specific CLIP teacher, where we can switch it to off-the-shelf SoTA CLIP models with different model scales flexibly to achieve better performance. For example, we can simply change the CLIP source from OpenAI-CLIP \citep{radford2021learning}, OpenCLIP \citep{cherti2023reproducible} to the best EVA-CLIP \citep{sun2023eva}, and probably to the better CLIP in the future. We can also directly scale up the CLIP teacher from EVA-CLIP-B (150 M) to EVA-CLIP-E (5 B). This demonstrates the flexibility and scalability of \Ours and shows the potential of \Ours to progress with the progress of CLIP models.

\begin{table}[t]
\scriptsize
  \setlength{\tabcolsep}{5pt}
  \centering
  \caption{Zero-shot classification on Objaverse-LVIS~\citep{deitke2023objaverse}, ModelNet40~\citep{wu20153d}, and ScanObjectNN~\citep{uy2019revisiting}. ($\dagger$ represents the best results achieved on different benchmarks respectively)}
  
  \resizebox{0.92\textwidth}{!}{
    \begin{tabular}{c|c|ccc|ccc|ccc}
    \toprule
    \multirow{2}[4]{*}{Method} & training shape  & \multicolumn{3}{c|}{Objaverse-LVIS} & \multicolumn{3}{c|}{ModelNet40} & \multicolumn{3}{c}{ScanObjectNN} \\
\cmidrule{3-11}          & source & Top1  & Top3  & Top5  & Top1  & Top3  & Top5  & Top1  & Top3  & Top5 \\
    \midrule
    ULIP-PointBERT & \multirow{3}[2]{*}{Ensembled}  & 21.4  & 38.1      & 46.0      & 71.4  & 84.4      & 89.2      & 46.0     & 66.1      & 76.4 \\
    OpenShape-SparseConv  & \multirow{3}[2]{*}{(no LVIS)} & 37.0  &   58.4    & 66.9  & 82.6  &   95.0    & 97.5  &    54.9   &  76.8     &  87.0 \\
     OpenShape-PointBERT & & 39.1 & 60.8 & 68.9 & {85.3} & 96.2 & 97.4 & 47.2 & 72.4 & 84.7 \\
     \textbf{\Ours} & & \textbf{47.2} & \textbf{68.8} & \textbf{76.1} & \textbf{86.8} & \textbf{97.3} & \textbf{98.4} & \textbf{66.5} & \textbf{83.5} & \textbf{90.1}  \\
    \midrule
    ULIP-PointBERT & \multirow{4}[2]{*}{Ensembled} & 26.8  & 44.8      & 52.6      & 75.1  & 88.1      & 93.2      & 51.6      & 72.5      & 82.3 \\
    OpenShape-SparseConv  &       & 43.4 & 64.8 & 72.4 & 83.4 & 95.6 & 97.8 & {56.7} &   78.9    & 88.6 \\
    OpenShape-PointBERT & & {46.8} & {69.1} & {77.0} & 84.4 & {96.5} & {98.0} & 52.2 & {79.7} & {88.7} \\
     \textbf{\Ours} & & \textbf{53.5} & \textbf{75.5} & \textbf{82.0} & \textbf{87.3} & \textbf{98.1} & \textbf{99.2} & \textbf{63.9} & \textbf{84.9} & \textbf{91.7}  \\

     \textbf{\Ours$\dagger$} & & \textbf{55.3} & \textbf{76.7} & \textbf{82.9} & \textbf{88.2} & \textbf{98.4} & \textbf{99.3} & \textbf{65.3} & \textbf{85.5} & \textbf{92.7}  \\

    \bottomrule
    \end{tabular}
    }
  \label{tab:zero-cls}%
    \vspace{-0.5cm}
  
\end{table}%

\section{Experiment}

\subsection{Zero-Shot Shape Classification}

We first evaluate \Ours under the zero-shot shape classification task. We conduct experiments under three benchmarks: ModelNet \citep{wu20153d}, ScanObjNN \citep{uy2019revisiting} and Objaverse-LVIS \citep{deitke2023objaverse}. ModelNet and ScanObjNN are widely-used datasets which contains 15 and 40 common categories, respectively. The Objaverse-LVIS benchmark is an annotated and cleaned subset of Objaverse which contains 46,832 shapes of 1,156 LVIS categories.
We follow the settings of OpenShape \citep{liu2023openshape} to conduct evaluations. For Objaverse-LVIS, we use 10,000 sampled colored points as input. For ModelNet40, we utilize 10,000 sampled points without color as input. For ScanObjNN, the input is 2,048 sampled points without color from the OBJ\_ONLY version.
We compare \Ours with the previous SoTA methods in the zero-shot shape classification task, such as PointCLIP \citep{zhang2022pointclip}, PointCLIP V2 \citep{zhu2022pointclipv2}, ULIP \citep{xue2023ulip} and OpenShape \citep{liu2023openshape}. Note that PointCLIP and PointCLIP V2 directly project point clouds into images and leverage 2D CLIP for classification, while other methods adopt a similar schema to train a native 3D backbone for aligning 3D representations with image and text representations produced by a pretrained CLIP. We follow OpenShape \citep{liu2023openshape} to report the performance under two different training settings. ``Ensembled" indicates that the backbones are trained under all the four datasets same as OpenShape and ``Ensembled (no LVIS)" further excludes the shapes from the Objaverse-LVIS subset. We justify that even when LVIS shapes are included in the training shapes, i.e., the ``Ensembled" dataset, their test-time category labels are probably not included in the training texts. 
The quantitative comparison is shown in Tab. \ref{tab:zero-cls}, where \Ours significantly outperforms the previous state-of-the-art methods under different settings.

\subsection{Few-Shot Linear Probing}

\begin{wrapfigure}{r}{0.4\linewidth}
\label{fig:fewshot}
  \vspace{-0.7cm}

  \includegraphics[width=\linewidth]{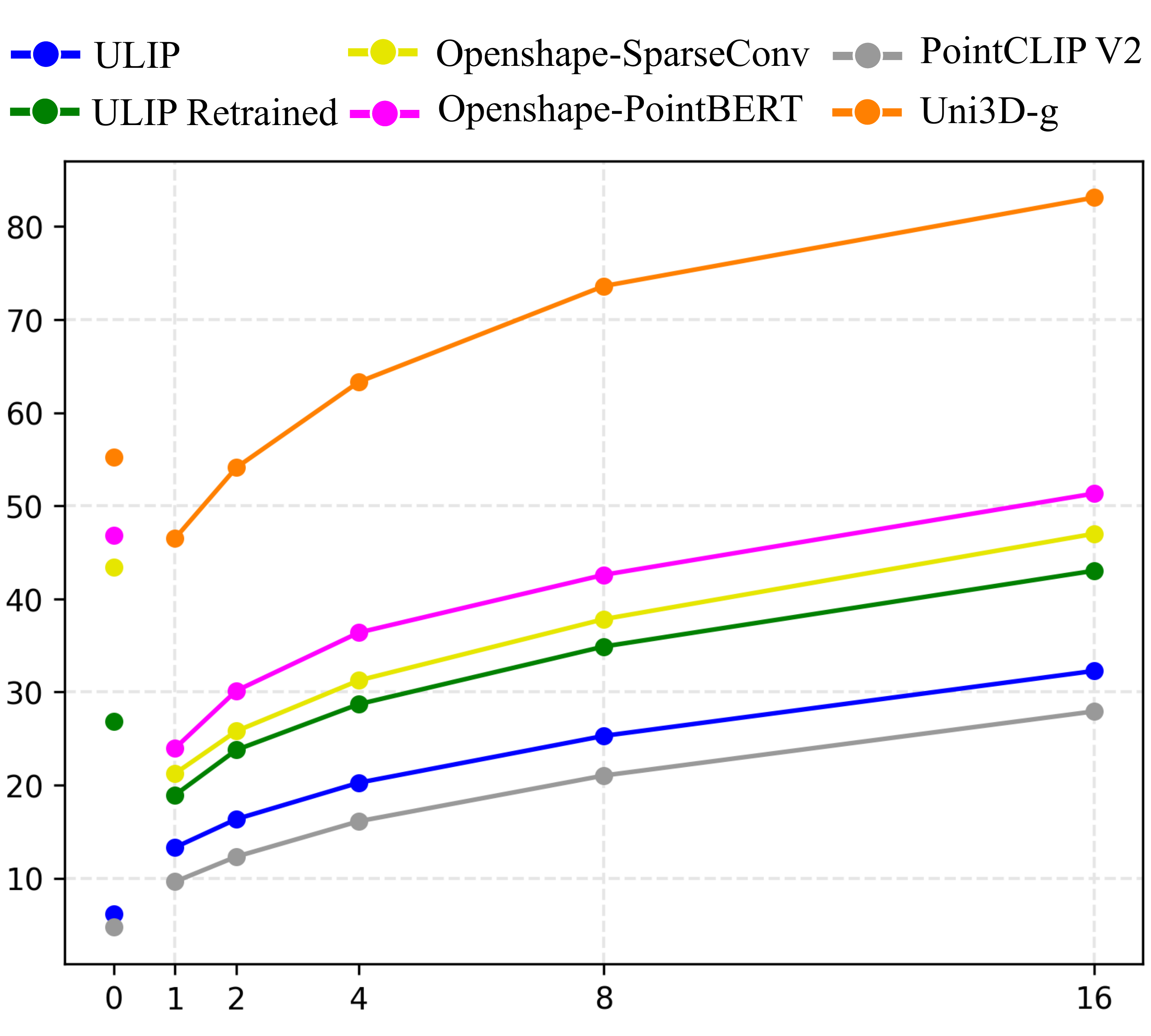}
  \vspace{-0.9cm}
  
  \caption{Few-shot linear probing on Objaverse-LVIS. We report the average performance over 10 random seeds.}
  \vspace{-0.5cm}
\end{wrapfigure}

Linear probing is a widely used approach for evaluating the learned representation of a model. To evaluate the linear probing ability of \Ours, we follow the common setting as OpenShape \citep{liu2023openshape} to freeze the parameters of \Ours and only train a linear classifier on few-shot class labels. We conduct few-shot linear probing under the difficult Objaverse-LVIS dataset with labeled training samples per class from 1, 2, 4, 8 to 16. Fig. \hyperref[fig:fewshot]{4} summarizes the performance of \Ours in comparison with OpenShape \citep{liu2023openshape} (PointBERT backbone and SparseConv backbone), ULIP \citep{xue2023ulip} (official release and the version retrained on the large ensembled dataset) and PointCLIP V2 \citep{zhu2022pointclipv2}. \Ours significantly outperforms all the other methods by a large margin under all the few-shot settings.

\subsection{Open-World Understanding}

To evaluate the capability of \Ours in 3D understanding of real-world shapes and scenes, we follow CLIP$^2$ \citep{zeng2023clip2} to conduct experiments under ScanNet \citep{dai2017scannet} to explore the zero-shot recognition performance of \Ours under real-world scenarios. Note that the ground truth instant segmentation is available for all the methods and the target is to recognize the category of each instant of the scene in a zero-shot way. ScanNet \citep{dai2017scannet} is a popular real-scanned 3D dataset containing 1.5K reconstructed meshes of real-world scenes. We adopt the same setting as CLIP$^2$ to split classes and evaluate the results under the test set of ScanNet. 

\begin{figure}
    \centering
    \includegraphics[width=\textwidth]{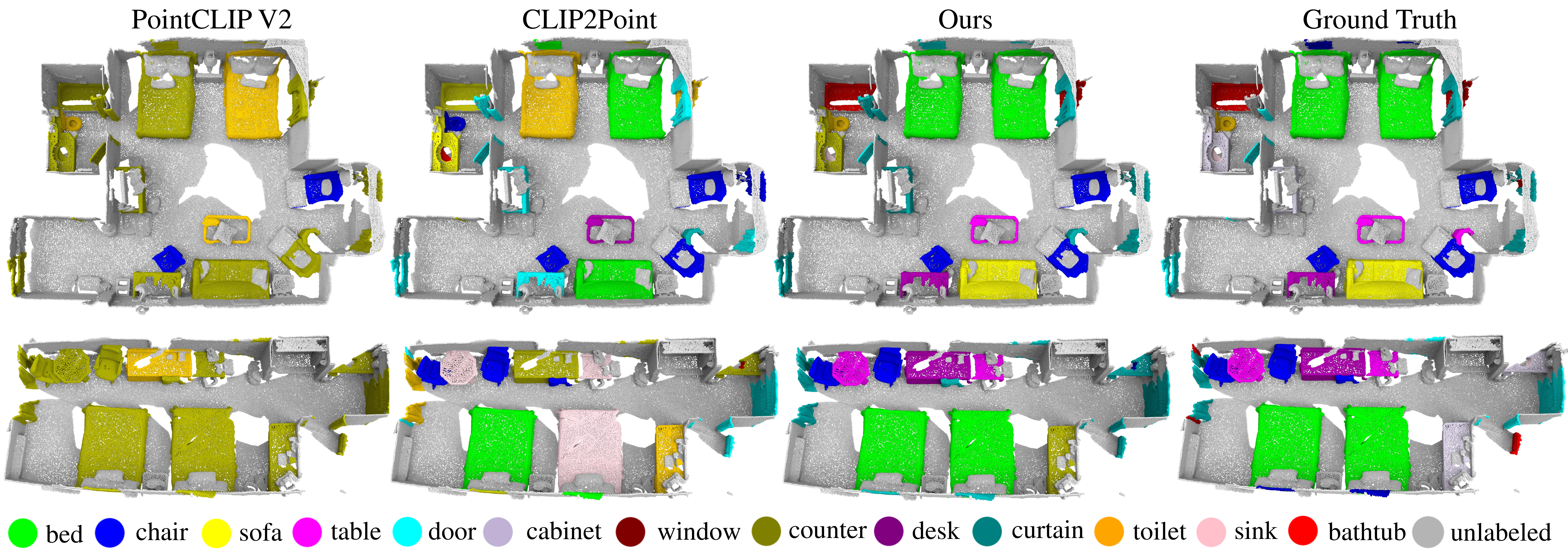}
    \vspace{-0.75cm}
    \caption{Comparisons of real-world zero-shot recognition results on Scannet dataset.}
    \vspace{-0.45cm}
    \label{fig:scannet}
\end{figure}

We compare our proposed \Ours with the state-of-the-art methods PointCLIP \citep{zhang2022pointclip}, PointCLIP V2 \citep{zhu2022pointclipv2}, CLIP2Point \citep{huang2022clip2point} and CLIP$^2$ \citep{zeng2023clip2}. The quantitative comparison is shown in Tab. \hyperref[tab:scannet]{2}. ``PointCLIP w/TP" and ``CLIP2Point w/TP'' donate training PointCLIP and CLIP2Point with the real-world data provided by CLIP$^2$. Note that ``PointCLIP w/TP", ``CLIP2Point w/TP" and CLIP$^2$ are trained under 1.6M triplets of real-world point cloud-image-text samples, while \Ours is only trained under available synthetic data. Nonetheless, \Ours achieves the best performance among all the previous methods. The results demonstrate the capability of \Ours to perform real-world recognition and understanding even without training under real-world data. The reason is that \Ours distills some perceptions of the real world from the CLIP models which are trained under large-scale real-world images and text. Moreover, by scaling up model size, \Ours achieves a larger representation bandwidth, leading to superior performance under difficult real-world rscenarios. The qualitative comparison is shown in Fig. \ref{fig:scannet}, where \Ours produces much more accurate zero-shot recognition results than PointCLIP V2 and CLIP2Point. We do not visually compare with CLIP$^2$ since its code and model are not publicly available.

\begin{table}[t]
\begin{center}
    \caption{Zero-shot recognition in ScanNet. Avg.: the average Top1 accuracy across all categories. 
}
\vspace{-0.25cm}
\resizebox{\textwidth}{!}{
\begin{centering}
\begin{tabular}{c|c|ccccccccccccccccc}
\toprule

Method & Avg.  & Bed & Cab & Chair & Sofa & Tabl & Door & Wind & Bksf & Pic & Cntr & Desk& Curt & Fridg & Bath & Showr & Toil & Sink  \tabularnewline
\midrule
PointCLIP & 6.3 & 0.0  & 0.0 & 0.0 &0.0  & 0.7 &0.0  & 0.0 & 91.8 & 0.0 &0.0 &0.0  &15.0  &0.0  &0.0 &0.0 &0.0 &0.0  \tabularnewline
PointCLIP V2 & 11.0 & 0.0  & 0.0 & 23.8 &0.0  & 0.0 &0.0  & 0.0 & 7.8 & 0.0 &90.7 &0.0  &0.0  &0.0  &0.0 &64.4 &0.0 &0.0  \tabularnewline
CLIP2Point & 24.9  & 20.8 & 0.0 & 85.1 & 43.3 & 26.5 &  69.9 & 0.0& 20.9 & 1.7 & 31.7 & 27.0 & 0.0 & 1.6 & 46.5&0.0&22.4&25.6    \tabularnewline
PointCLIP w/  TP. &26.1&0.0&55.7&72.8&5.0&5.1&1.7&0.0&77.2&0.0&0.0&51.7&0.3&0.0&0.0&40.3&85.3&49.2\tabularnewline
CLIP2Point w/  TP. & 35.2  &11.8&3.0&45.1&27.6&10.5&61.5&2.6&71.9&0.3&33.6&29.9&4.7& 11.5&72.2&92.4&86.1&34.0\tabularnewline
CLIP${^2}$ & 38.5 & 32.6& {67.2}& 69.3 & 42.3 & 18.3 & 19.1 & 4.0 & 62.6 & 1.4 & 12.7 & 52.8 & 40.1 & 9.1 & 59.7 & 41.0 & 71.0 & 45.5  \tabularnewline
\midrule
\textbf{Uni3D} & \textbf{45.8} & {58.5}& 3.7& {78.8} & {83.7} & {54.9} & 31.3 & {39.4} & 70.1 & {35.1} & 1.9 & {27.3} & {94.2} & 13.8 & 38.7 & 10.7 & 88.1 & {47.6}  \tabularnewline
\bottomrule
\end{tabular}
\end{centering}
		}
	\end{center}

	\vspace{-0.65cm}
\label{tab:scannet}
\end{table}

\subsection{Open-Vocabulary / Few-Shot Part Segmentation}

Some prior methods \citep{rao2022denseclip,yang2022lavt} have demonstrated that transferring the knowledge gained from image-text contrastive learning, i.e., CLIP, can yield significant performance improvements in 2D dense prediction tasks (e.g. segmentation and detection). However, transferring this knowledge to 3D dense prediction tasks is barely explored. We propose a novel approach for 3D dense prediction with \Ours, and justify the effectiveness with part segmentation experiment. For more details on the approach, please refer to Sec. \ref{sec.appendP} of the appendix.

\begin{wraptable}{r}{0.45\columnwidth}\small
\vspace{-0.7cm}
\centering
   \caption{Few-shot part segmentation results on the ShapeNetPart dataset.}
\resizebox{0.45\columnwidth}{!}{
   
   \begin{tabular}{c|c|c|c|c}
    \toprule
     Method&Data&mIoU$_C$&Data&mIoU$_C$ \\
      \midrule
      PointNet&\multirow{5}{*}{{\thead{10\% \\ train set}}}&72.7&\multirow{5}{*}{{\thead{20\% \\ train set}}}&73.5\\
      PointNet++&&74.8&&76.8\\
      PointCNN&&60.4&&64.1\\
      SSCN&&60.2&&65.2\\
      PointBERT&&76.4&&79.6\\
      \midrule
      PointBERT&\multirow{2}{*}{{1-shot}}&66.2&\multirow{2}{*}{{2-shot}}&71.9 \\
      
      \textbf{Uni3D}&&\textbf{75.9}&&\textbf{78.2}\\
    \bottomrule
   \vspace{0.85cm}
   \end{tabular}}
   \label{table:fewshotpart}
   \vspace{-1.3cm}
\end{wraptable}

We conduct part segmentation experiments under ShapeNetPart dataset \citep{yi2016scalable}. The results in Tab. \ref{table:fewshotpart} demonstrate that when supervised with only 1 or 2 samples per class, \Ours outperforms PointBERT by +13.3\%/+9.8\%. Moreover, we largely increase the training samples used for comparative methods to 10\% or 20\% of the training set. These settings surpass training samples in \Ours's one-shot or two-shot settings by two orders of magnitude. Even in the face of such a discrepancy in the number of training samples, \Ours still achieves comparable performance in terms of overall mIoU. 
The visual comparisons with PointBERT are provided in Sec.~\ref{sec.appendP} of the appendix.

\begin{table*}[t]
\setlength{\tabcolsep}{1.5pt}
\centering
   \caption{Open-vocabulary segmentation results on the ShapeNetPart dataset. }
\vspace{0.2cm}
   
\resizebox{\linewidth}{!}{
    \begin{tabular}{c|c|cccccccccccccccccc}

      \toprule
      \multirow{2}{*}{{Method}}&\multicolumn{11}{c|}{Seen Categories}&\multicolumn{7}{c|}{Unseen Categories}&\\
     \cline{2-20}
     	&mIoU$_C$&Car&Knife&Lamp&Moto&Pistol&Rocket&Guitar&Skate&Chair&\multicolumn{1}{c|}{Cap}  &Plane&Bag&Earph&Laptop&Mug&\multicolumn{1}{c|}{Table}  &\multicolumn{1}{c|}{mIoU$_C$} &mIoU$_C$-ALL \\
     \midrule
     \textbf{Uni3D}&\textbf{76.0}&74.1&86.3&85.7&62.7&77.8&41.7&89.3&71.0&89.9&\multicolumn{1}{c|}{81.8}&23.4&57.1&57.1&26.2&50.0&\multicolumn{1}{c|}{54.4}&\multicolumn{1}{c|}{\textbf{44.7}}&\textbf{64.3}\\

\hline

   \hline
   \end{tabular}}
   \label{table:openpart}
\end{table*}

``Open-vocabulary part segmentation'' quantifies the ability of \Ours to learn fine-grained semantic information of local point clouds during multi-modal contrastive pre-training. We partition the ShapeNetPart dataset into two subsets: ``Seen Categories" and ``Unseen Categories.'' In the ``Seen Categories'' subset, the text of ground-truth part labels serve as training samples of \Ours for learning part semantics, while in the ``Unseen Categories'' subset, the text of ground-truth part labels is unseen during training and is only utilized for testing. The superior performance of \Ours in Tab.~\ref{table:openpart} demonstrates its ability to discern fine-grained 3D patterns, even for part-level semantic concepts not encountered in the ``Seen Categories''. These results robustly affirm \Ours's capacity to transfer the learned patterns in a close set of 3D parts to open-vocabulary parts, utilizing the rich open-world knowledge distilled from the pre-trained CLIP model. We believe that \Ours opens avenues to achieve fine-grained, cross-category segmentation of open-vocabulary 3D concepts by leveraging a limited number of category-agnostic segmentation examples.

\subsection{Point Cloud Painting}

\begin{figure}[tb]
    \centering
    \includegraphics[width=\textwidth]{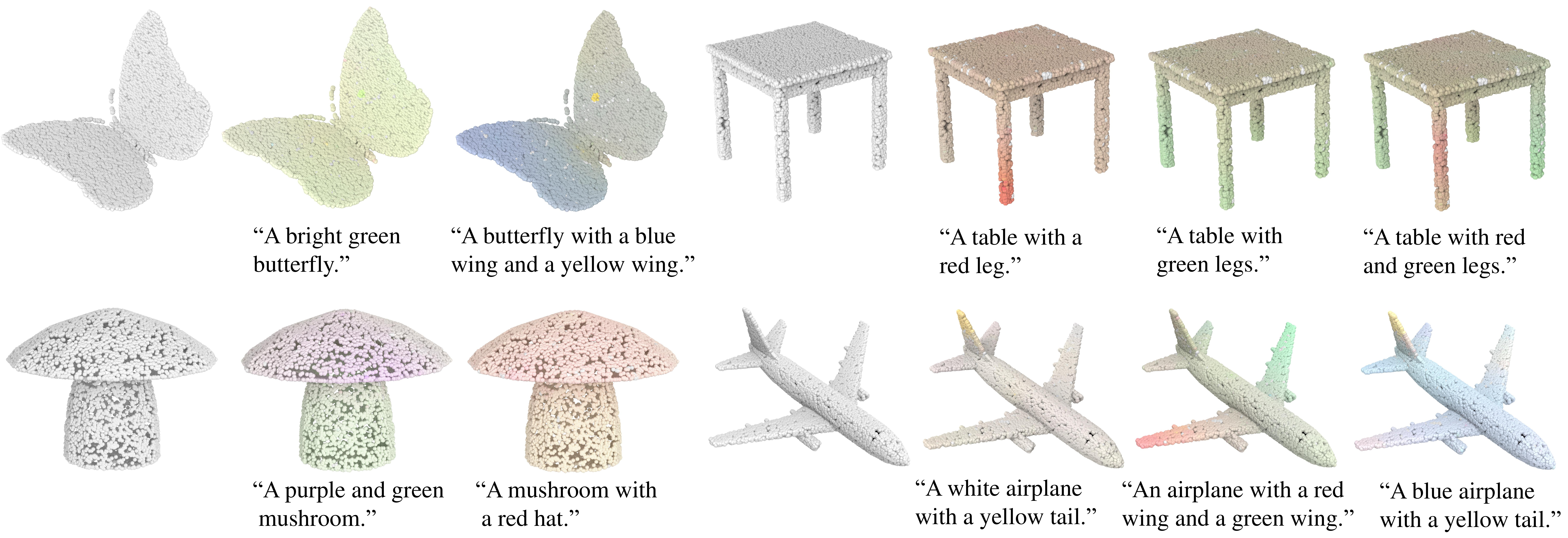}
    \vspace{-0.3cm}
    \caption{Point cloud painting results. The white models are the initial point clouds.}
    \label{fig:editing}
    \vspace{-0.2cm}
\end{figure}

We propose to leverage the trained \Ours for painting point clouds by exploring the learned 3D semantic patterns in \Ours. Specifically, given an initial point cloud and an input prompt, we optimize the appearance, i.e., RGB channel of the point cloud, by maximizing the cosine similarity between the feature of the point cloud extracted by Uni3D and the feature of the prompt extracted with CLIP text encoder. The painting for a point cloud can be achieved within one minute in a single V100 GPU. We show the paintings in Fig. \ref{fig:editing}, where \Ours successfully optimizes the point cloud by revealing complex semantics from the prompt. The results demonstrate that \Ours has learned abundant and diverse 3D patterns via contrastive pretraining.

\subsection{Cross-Modal Retrieval}
\begin{figure}[tb]
    \centering
    \includegraphics[width=\textwidth]{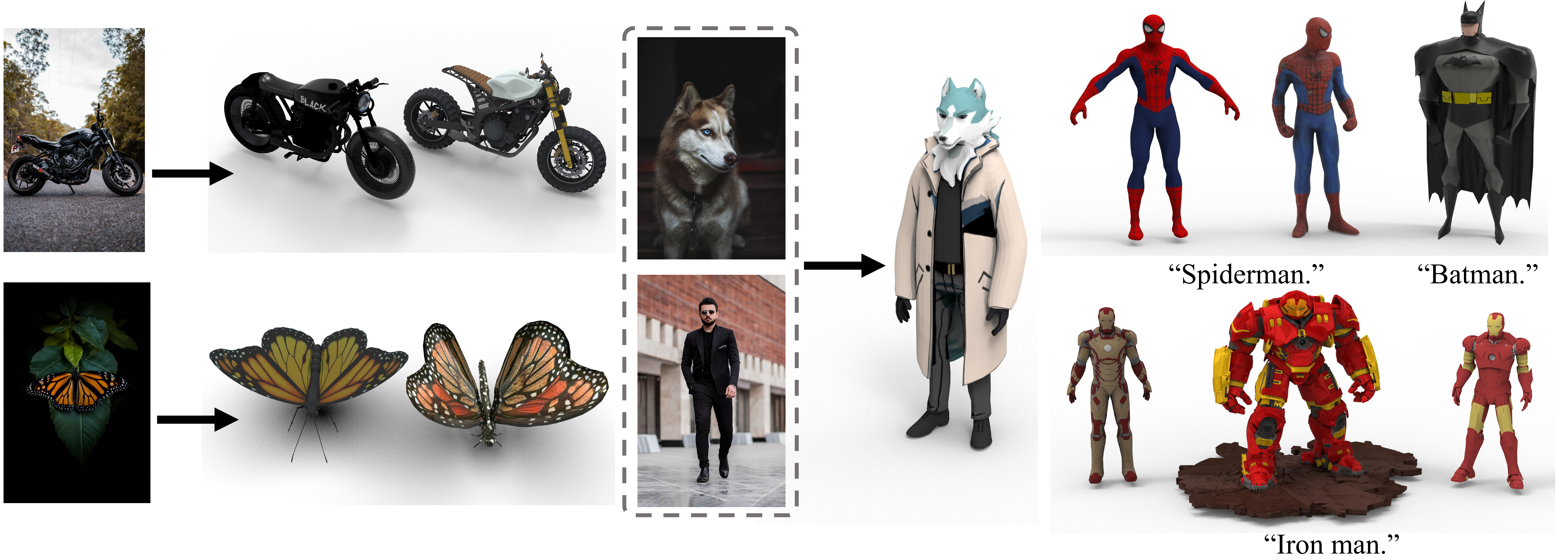}
    \vspace{-0.5cm}
    \caption{Image-query/text-query 3D shape retrieval results. In the first column, we query two most similar 3D shapes for each query image. In the second column, we  take two images as inputs and retrieve the 3D shape similar to both two images. In the third column, we query three most similar 3D shapes for each query text.}
    \label{fig:retrival_image_text}
    \vspace{-0.5cm}
\end{figure}

With the learned multi-modal representations of \Ours, we can naturally retrieve 3D shapes from images or text. Specifically, we retrieve 3D shapes from the large 3D dataset \citep{deitke2023objaverse} by calculating the cosine similarity between the embedding of a query image or a query text prompt and the embedding of 3D shapes. We then perform kNN to achieve the most similar 3D shapes of the query. In Fig. \ref{fig:retrival_image_text}, we show that \Ours successfully retrieves 3D shapes from real-world images. Note that the images for training are only renderings, and there is a big gap between the training images and the real-world images. We also take two images as inputs and retrieve the shape similar to both two images by calculating the cosine similarity between the average of the embedding of two images and the embedding of 3D shapes. The interesting results demonstrate that \Ours learns a diverse 3D representation with the ability to perceive multiple 2D signals. We further show the results of leveraging \Ours to retrieve 3D shapes from the input texts in Fig. \ref{fig:retrival_image_text}. More visualization results are provided in Sec.~\ref{sec.appendR} of the appendix.

\subsection{Ablation Study}

We then conduct ablation studies to justify the effectiveness of each design in \Ours. The default setting is to use the ViT-Base as the backbone with an initialization of EVA \citep{fang2023eva}, and the default CLIP teacher is EVA-CLIP-E \citep{sun2023eva}. The default data setting is ``Ensembled (no-LVIS)". We keep the default experimental setting during ablation studies except for the modified part described in each ablation experiment below.

\textbf{Scaling Up Model Size.}
We first explore the effectiveness of scaling up the model size of \Ours in Tab.~\ref{tab:model_scaling}. Since we leverage a unified vanilla transformer structurally equivalent to ViT as the foundational 3D representation model, we can simply scale up \Ours with the well-studied unified 2D/NLP scaling-up strategies. Specifically, we follow the scaling up principles of the plain ViT \citep{dosovitskiy2020image} to increase parameters from 6 M (Tiny), 23 M (Small), 88 M (Base), 307 M (Large) to 1 B (giant). The hyper-parameters on the model architecture are detailed in Tab. \ref{tab:model_scaling}. The performance under different model scales demonstrates that scaling up the model size of \Ours can significantly improve the 3D representation.

\begin{table}[tb]
    \centering
        \caption{Scaling up model size in Uni3D. $\diamond$ represents the results under ``Ensembled'' dataset without LVIS shapes. $\dagger$ represents the results under ``Ensembled'' dataset with LVIS shapes.
}
\vspace{0.2cm}

    \begin{tabular}{l|ccc|c|cc|cc}
        \toprule
        Model & Depth & Width & Heads & \#Params & MNet40$^\diamond$ & O-LVIS$^\diamond$ & MNet40$^\dagger$ & O-LVIS$^\dagger$ \\ \midrule
        Uni3D-Ti & 12  & 192 & 3 & 6.2M & 85.8 & 43.5 & 85.9 & 46.5 \\
        Uni3D-S & 12  & 384 & 6 & 22.6M & 86.0 & 44.8 & 86.0 & 50.6 \\
        Uni3D-B & 12 & 768 & 12 & 88.4M & 86.2 & 45.8 & 86.5 & 51.6\\
        Uni3D-L & 24 & 1024 & 16 & 306.7M & 86.6 & 46.2 & 86.6 & 53.2 \\             Uni3D-g & 40 & 1408 & 16 & 1016.5M & 86.8 & 47.2 & 88.2 & 55.3 \\
        \bottomrule
    \end{tabular}

    \label{tab:model_scaling}
    \vspace{-0.6cm}
\end{table}

\textbf{Switching / Scaling Up CLIP Teachers.}
We justify that \Ours is a flexible framework where we can switch the off-the-shelf SoTA CLIP models as the teacher. To this end, we investigate the performances of \Ours with different CLIP teachers at different scales. Specifically, we evaluate various CLIP models (e.g. OpenAI-CLIP \citep{radford2021learning}, OpenCLIP \citep{cherti2023reproducible} and EVA-CLIP \citep{sun2023eva}), and also explore large scale CLIP models (e.g., OpenCLIP-bigG, EVA-CLIP-E). The quantitative comparison is shown in Tab. \ref{tab:teacher_scaling}, with the best performance achieved by the largest CLIP model EVA-CLIP-E. The results show that the capability and model scale of CLIP teachers are key factors for achieving better performance in \Ours. Moreover, it indicates the potential of \Ours to progress with the progress of CLIP models by switching state-of-the-art CLIP teachers.

\begin{table}
  \begin{minipage}{0.60\textwidth}
    \caption{Different CLIP teachers at different model scales.}
    \vspace{0.3cm}
    \centering
    \begin{tabular}{lccc}
        \toprule
        CLIP variant & Pretrain data & \#Params & O-LVIS \\ \hline
        EVA-CLIP-B/16 & Merged-2B & 150M &42.3  \\
        OpenAI-CLIP-B/16 & WIT-400M & 150M & 42.7 \\
        OpenCLIP-B/16 & LAION-2B & 150M & 43.4 \\ 
        \midrule
        OpenCLIP-bigG/14 & LAION-2B & 2.5B & 44.5 \\
        EVA-CLIP-E/14+ & LAION-2B & 5.0B & 45.8 \\
        \bottomrule
    \end{tabular}
    \label{tab:teacher_scaling}
    
  \end{minipage}%
  \hspace{0.5cm}
  \begin{minipage}{0.35\textwidth}
    \caption{Initializing Uni3D with different pretrained large models.}
\vspace{0.2cm}
    
    \begin{tabular}{l|c}
        \toprule
        Init variant &  O-LVIS \\ \midrule
        None & 44.8 \\
        DINO & 45.0\\
        EVA-CLIP & 45.2 \\
        EVA & 45.8  \\
        \midrule
        EVA + Freeze ViT & 15.7  \\

        \bottomrule
    \end{tabular}

    \label{tab:init_vit1}
  \end{minipage}
  \vspace{-0.3cm}
\end{table}

\textbf{Initializing Transformer.} 
\label{sec.init}
We further conduct ablation studies to explore the effectiveness of initializing \Ours with 2D pretraining or multi-modal large models. In Tab. \ref{tab:init_vit1}, we report the performance of training \Ours from scratch (None) and initializing \Ours with off-the-shelf 2D pretraining model DINO \citep{caron2021emerging} / EVA \citep{fang2023eva} and SoTA CLIP model EVA-CLIP \citep{sun2023eva}. The best performance is achieved by initializing \Ours with the SoTA 2D pretraining model EVA \citep{fang2023eva}. We also demonstrate that leveraging the frozen parameters from the 2D pretrained ViT model may fail to provide strong 3D understanding without fine-tuning, as shown in ``EVA + Freeze ViT" of Tab. \ref{tab:init_vit1}. For more analysis on initializing \Ours, please refer to Sec. \ref{appendix:init} of the appendix.

\section{Conclusion}
We present \Ours, a unified framework that scales up a 3D representation model to one billion parameters. We directly leverage a unified vanilla transformer structurally equivalent to ViT as the model, which allows us to simply scale up \Ours with the well-studied unified 2D/NLP scaling-up strategies. Moreover, Uni3D can leverage abundant 2D pretrained models as initialization and image-text aligned models as the target, unlocking the great potential of 2D models and strategies to the 3D world. We train \Ours under a large dataset containing about one million 3D point clouds, 10 million images and 70 million texts to explore powerful 3D representations by aligning the 3D point cloud features with the image-text aligned features.  \Ours achieves state-of-the-art performance in various 3D understanding tasks including zero-shot and few-shot classification, open-world understanding, part segmentation, etc.
We believe that \Ours can serve as a 3D foundation model to enable many applications in the 3D community.

\bibliography{iclr2024_conference}
\bibliographystyle{iclr2024_conference}

\clearpage
\appendix
\section{Training Details}
\label{sec.m4}
We freeze the CLIP text and image encoders while focusing on training the 3D encoder utilizing the cross-modal contrastive loss. We employ the Adam~\citep{kingma2014adam} optimizer with a peak learning rate of 1e-3 that gradually decreases following a cosine learning rate schedule. To enhance training stability, we adopt stochastic depth~\citep{huang2016deep} regularization.
We also leverage the FLIP~\citep{li2023scaling} technique, which randomly masks 50\% of point tokens during training, reducing time complexity by half. 
We precache text and image CLIP embeddings of all shapes, allowing us to increase the total batch size to 1152 and greatly accelerating training. To further improve the training process, we adopt DeepSpeed~\citep{rasley2020deepspeed} with ZeRO stage-1 optimizer and fp16 precision with dynamic loss scaling~\citep{rajbhandari2020zero}. 
Taking advantage of the aforementioned strategies, our largest model, i.e., Uni3D-g with one billion parameters, converges in approximately 20 hours with 24 $\times$ NVIDIA-A100-SXM4-40GB GPUs.

\section{Part Setmentation Details}
\label{sec.appendP}
Some prior methods \citep{rao2022denseclip,yang2022lavt} have demonstrated that transferring the knowledge gained from image-text contrastive learning, i.e., CLIP, can yield significant performance improvements in 2D dense prediction tasks (e.g. segmentation and detection). However, transferring this knowledge to 3D dense prediction tasks is barely explored. We seek to find a way to convert the global point cloud-text alignment learned by \Ours into a local point-text alignment. We aim to demonstrate that the object-level  pre-training in \Ours is sufficient for learning detailed local 3D visual concepts. 
Specifically, we select the features from $4^{th}$, $8^{th}$ and the last layer of the ViT in \Ours, denoted as $H^{4}$, $H^{8}$ and $H^{12}$. Following PointNet++~\citep{qi2017pointnet++}, we employ feature propagation to upsample group features $H^{4}$, $H^{8}$ and $H^{12}$ into point-wise features. During training, we freeze the \Ours backbone and only optimize the parameters in the feature propagation layer, with supervision to align point-wise features and text features of ground-truth part labels, which are extracted by the CLIP text encoder. 
By freezing the parameters of learned \Ours, we focus on effectively exploring the pre-trained fine-grained knowledge.

\begin{figure}[b]
    \centering
    \includegraphics[width=\textwidth]{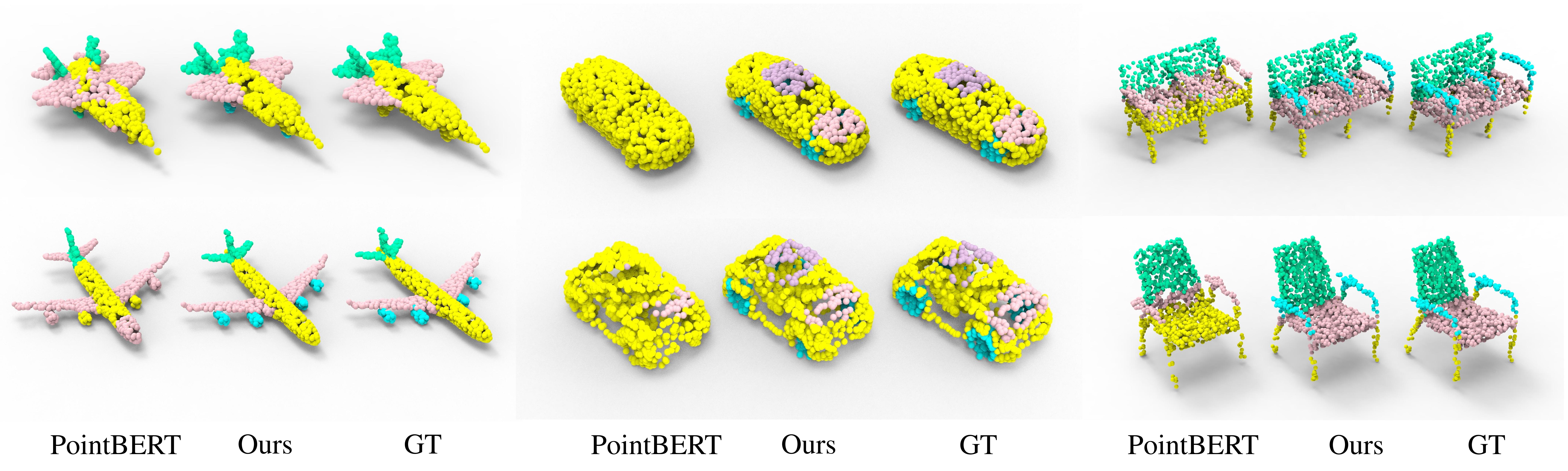}
    \caption{Comparison in one-shot part segmentation under ShapeNetPart.}
    \label{fig:part2}
\end{figure}

The visual comparison in Fig. \ref{fig:part2} shows that our method can produce more accurate segmentation results in the one-shot part segmentation setting. 

\section{More Visualization of Cross-Modal Retrieval}
\label{sec.appendR}

\begin{figure}[tb]
    \centering
    \includegraphics[width=\textwidth]{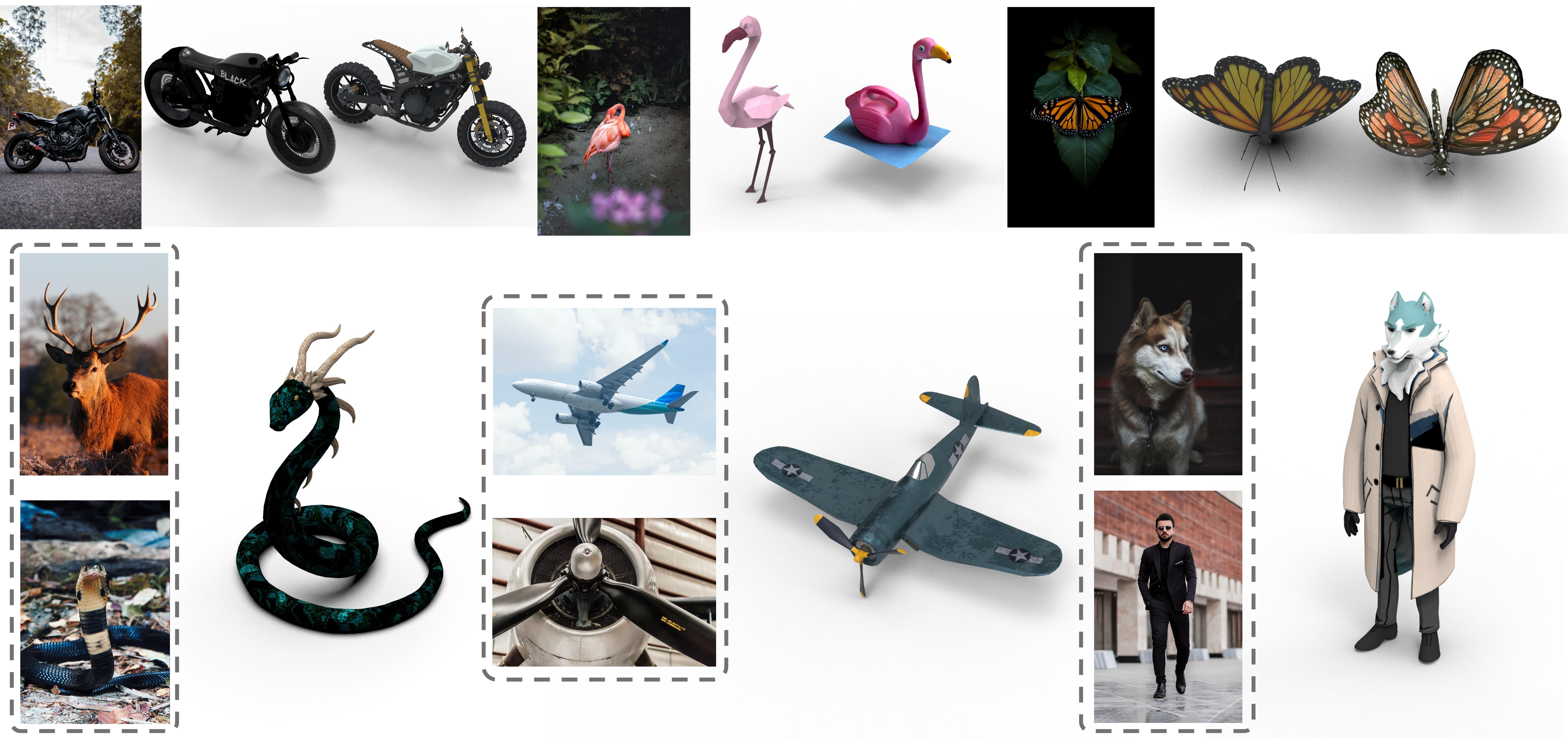}
    \caption{Image-query 3D shape retrieval results. In the first row, we query two most similar 3D shapes for each query image. In the second row, we  take two images as inputs and retrieve the 3D shape similar to both two images.}
    \label{fig:retrival_image}
\end{figure}

\begin{figure}[tb]
    \centering
    \includegraphics[width=\textwidth]{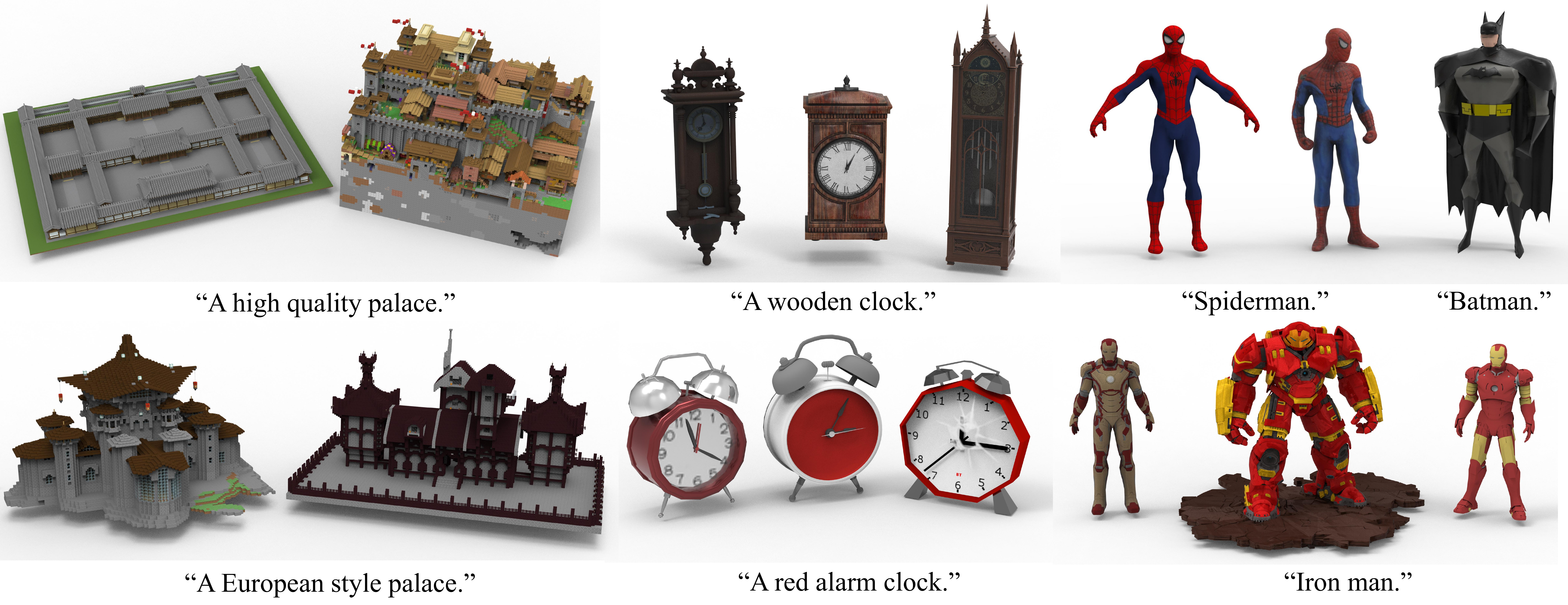}
    \caption{Text-query 3D shape retrieval results.}
    \label{fig:retrival_text}
\end{figure}

In Fig. \ref{fig:retrival_image}, we visualize more 3D shapes retrieved from real-world one or multiple images.  We further show the results of leveraging \Ours to retrieve 3D shapes from the input texts in Fig. \ref{fig:retrival_text}.

\section{More Analysis on Initializing \Ours}
\label{appendix:init}

As demonstrated in Tab. \ref{tab:init_vit1}, the best performance is achieved by initializing \Ours with the SoTA 2D pretraining model EVA \citep{fang2023eva}. The reason is that EVA model learns powerful and general representations to serve as a fine initialization of cross-modal contrastive learning (e.g. CLIP) as demonstrated in EVA \citep{fang2023eva}, EVA-02 \citep{fang2023eva02} and EVA-CLIP \citep{sun2023eva}. 

\begin{wraptable}{r}{0.45\columnwidth}\small
\vspace{-0.8cm}

    \centering
    \caption{Effect of initializing \Ours under two-modal situation.}
\vspace{0.1cm}
    
    \begin{tabular}{l|c}
        \toprule
        Init variant &  O-LVIS \\ \midrule
        None Init + only text & 20.7  \\
        None Init + only image & 12.4  \\
        EVA + only text & 40.1  \\
        EVA + only image & 26.3  \\
        \bottomrule
    \end{tabular}
\vspace{-0.6cm}

    \label{tab:init_vit}
\end{wraptable}
We justify that \Ours also learns a cross-modal representation similar to CLIP, where the general patterns learned by EVA play a key role in improving and stabilizing the training of \Ours.  The analysis is further supported by the results of two-modal contrastive learning as shown in Tab. \ref{tab:init_vit}. Specifically, we conduct experiments to train \Ours with only contrastive loss with CLIP image features (+only image) or CLIP text features (+only text), respectively. The results show that the optimization crashed without EVA initialization in the difficult situation where only images are available (20.7 vs.\ 40.1) or only texts are available (12.4 vs.\ 26.3).

\section{Few-shot results}
We conduct few-shot linear probing under the difficult Objaverse-LVIS dataset with labeled training samples per class from 1, 2, 4, 8 to 16. The comparison is shown in Fig. \hyperref[fig:fewshot]{4}. We further provide the detailed quantitative results in Tab. \ref{tab:fewshot}. 

\begin{table}[t]
\centering
    \caption{Few-shot linear probing on Objaverse-LVIS. We report the average Top1 accuracy over 10 random seeds. 
}
\begin{tabular}{c|cccccccccccccccccc}
\toprule

Objaverse-LVIS & 1-shot & 2-shot & 4-shot & 8-shot & 16-shot \\
\midrule
PointCLIP V2 & 9.6&12.3&16.1&21.0&27.9\\
ULIP & 13.3& 16.3& 20.3& 25.3& 32.3 \\
ULIP Retrained & 18.9& 23.8& 28.7&34.9&43.0\\
OpenShape-PointBERT & 24.0& 30.1& 36.4& 42.6& 51.3 \\ 
OpenShape-SparseConv & 21.2& 25.8& 31.2& 37.8& 47.0 \\ 
\textbf{\Ours} & \textbf{46.5}& \textbf{54.1}& \textbf{63.3}& \textbf{73.6}& \textbf{83.1} \\

\bottomrule
\end{tabular}

\label{tab:fewshot}
\end{table}

\end{document}